%% file: lrec2016-t4f.tex
\documentclass[10pt, a4paper]{article}
\usepackage{lrec2016}
\usepackage{graphicx}
\usepackage{url}
\usepackage{latexsym}
\usepackage{booktabs}
\usepackage{tikz}
\usepackage{color}
\usepackage{caption}
\usepackage{subcaption}

\DeclareGraphicsExtensions{.pdf,.png,.jpg}

\title{Towards Using Social Media to Identify Individuals at Risk for Preventable Chronic Illness}

\name{Dane Bell, Daniel Fried, Luwen Huangfu, Mihai Surdeanu, Stephen Kobourov}

\address{ University of Arizona \\
               Tucson, AZ 85721, USA \\
               \{dane, dfried, huangfuluwen, msurdeanu, kobourov\}@email.arizona.edu\\}

\abstract{
We describe a strategy for the acquisition of training data necessary to build a social-media-driven early detection system for individuals at risk for (preventable) type 2 diabetes mellitus (T2DM). The strategy uses a game-like quiz with data and questions acquired semi-automatically from Twitter. The questions are designed to inspire participant engagement and 
collect relevant data to train a public-health model applied to individuals.
Prior systems designed to use social media such as Twitter to predict obesity (a risk factor for T2DM) operate on entire communities such as states, counties, or cities, based on statistics gathered by government agencies. Because there is considerable variation among individuals within these groups, training data on the individual level would be more effective, but this data is difficult to acquire.
The approach proposed here aims to address this issue.
Our strategy has two steps. First, we trained a random forest classifier on data gathered from (public) Twitter statuses and state-level statistics with state-of-the-art accuracy. We then converted this classifier into a 20-questions-style quiz and made it available online. In doing so, we achieved high engagement with individuals that took the quiz, while also building a training set of voluntarily supplied individual-level data for future classification.
\\ \newline }

\begin{document}

\maketitleabstract

\input{intro}

\input{prior}

\input{method}

\input{results}

\input{discussion}
\input{conclusion}
\input{resources}

%\section*{Acknowledgments}
% Do not include acknowledgments when submitting paper for review.

\section{References}

\bibliographystyle{lrec2016}
\bibliography{t4f}

\end{document}

%% file: intro.tex
\section{Introduction}
 % Key points:
 % - Obesity is epidemic in the United States and increasingly elsewhere in the developed world.
 % - Any reduction in obesity rates would lead to massive healthcare savings, because obesity leads to many health problems
 % - Intervention by social media has been shown to have modest but significant success in decreasing obesity
 % - Automatic detection using machine learning is an extremely cost effective way to test for obesity
 % Mihai has a grant application with a lot of information and citations about social media health interventions.

Data collection in the public health domain is difficult due to privacy concerns and low engagement.
For example, people seldom engage with surveys that require them to report their height and weight. 
However, such data is crucial for training automated public health tools, such as algorithms that detect risk for (preventable) type 2 diabetes mellitus (T2DM, henceforth {\it diabetes}). We propose a semi-automated data collection algorithm for obesity detection that mitigates these issues with a game-like quiz that is automatically bootstrapped from a machine-learning model trained over social media data. The resulting quiz is non-intrusive, focusing on ``fun'' questions about food and food language while avoiding personal questions, which leads to high engagement.

We believe this idea contributes to addressing one of the most challenging unsolved public health problems: the high rate of chronic illness resulting from modifiable risk 
factors such as poor diet quality and physical inactivity. It is estimated that more than 86 million Americans over the age of 20 
exhibit signs of pre-diabetes, and 70\% of these pre-diabetic individuals will eventually develop T2DM, 
a chronic and debilitating disease associated with heart disease, stroke, blindness, kidney failure, and amputations~
\cite{cdc14diagnosed,american2008diagnosis}. 
In the United States, the estimated cost of T2DM rose to \$245 billion in 2012 \cite{american2013economic}.
Yet, 90\% of these individuals at high risk are not aware of it \cite{Li2013awareness}.

Our long-term goal is to develop tools that automatically classify overweight individuals (hence at risk for T2DM\footnote{In the CDC diabetes questionnaire available at \url{http://www.cdc.gov/diabetes/prevention/pdf/prediabetestest.pdf}, overweight BMI contributes more than half of the points associated with diabetes risk.}) using solely public 
social media information. 
The advantage of such an effort is that the resulting tool provides {\em non-intrusive} and {\em cost-effective} means 
to detect and warn at-risk individuals early, {\em before} they visit a doctor's office, and possibly influence their decision to visit a doctor.

Previous work has demonstrated that intervention by social media has modest but significant success in decreasing obesity~\cite{ashrafian:2014}. 
Furthermore, there is good evidence that detecting communities at risk using computational models trained 
on social media data is possible~\cite{fried:2014,culotta:2014}. However, in all cases, classification is made on aggregated data from cities, counties, or states, so these models are not immediately applicable to the task of classifying individuals. 

Our work takes the first steps towards transferring a classification model that identifies communities that are more overweight than average to classifying overweight (and thus at-risk for T2DM) {\em individuals}. The contributions of our work are:

%{\bf 1.} We improve the state-level classification approach of \newcite{fried:2014} with a random forest model that performs better 
%and is considerably more compact when trained on the same Twitter data, consisting of food-related tweets grouped by the state 
%of their users: using only 7 decision trees with a maximum depth of 3, our model predicts if a state is more/less overweight on 
%average than the median state with an accuracy of 82\%.

{\bf 1.} We introduce a random-forest (RF) model that classifies US states as more or less overweight than average using only 7 decision trees with a maximum depth of 3. Despite the model's simplicity, it outperforms \newcite{fried:2014}'s best model by 2\% accuracy.

{\bf 2.} Using this model, we introduce a novel semi-automated process that converts the decision nodes in the RF model into natural language questions. We then use these questions to implement a quiz that mimics a 20-questions-like game. The quiz aims to detect if the person taking it is overweight or not based on indirect questions related to food or use of food-related words. 
%The quiz is publicly available\footnote{\url{http://sites.google.com/site/twitter4food}}. 
To our knowledge, we are the first to use a semiautomatically generated quiz for data acquisition.

{\bf 3.} We demonstrate that this quiz serves as a non-intrusive and engaging data collection process for individuals\footnote{Previous work has demonstrated the high engagement of such quizzes. For example, the most popular post in the New York Times for 2013 was a quiz predicting respondents' locations by features of their dialect such as distinctive vocabulary: {\scriptsize \url{http://www.nytco.com/the-new-york-timess-most-visited-content-of-2013}}}. The survey was posted online and evaluated with 945 participants, of 
whom 926 voluntarily provided supplemental data, such as information necessary to compute the Body Mass Index (BMI), demographics, 
and Twitter handle, demonstrating excellent engagement. The random-forest model backing the survey agreed with self-reported BMI in 78.7\% of cases. More importantly, the differences prompted a spirited Reddit discussion, again supporting our hypothesis that this quiz leads to higher participant engagement\footnote{{\scriptsize  \url{http://www.reddit.com/r/SampleSize/comments/3hbiz3/academic_can_our_automatically_generated/}}}.

This initial experiment suggests that it is possible to use easy-to-access {\em community} data to acquire training data on {\em 
individuals}, which is much more expensive to obtain, yet is fundamental to building individualized public health tools.
The anonymized data collected from the quiz is publicly available.%\footnote{\url{http://git.io/ vZY5U}}. 
%\end{enumerate}

%% file: prior.tex
\section{Prior work}

% Cite first Twitter4Food paper, Fried et al. 2014, and briefly summarize relevant aspects of its construction, i.e. states were split
%on median obesity (and diabetes and political views) and localizable Twitter users' tweets were used to train several models to
%classify a held-out state as more or less obese (or diabetic or Democratic) than the median. Other work may be relevant as
%well.

% Cite Abbar et al. "You Tweet What You Eat" and talk about the assumption that an individual in an overweight county will be
%overweight. We want truly individual information, so this necessitates gathering individuals' information.

%
% ms: this is interesting, but not directly related to this paper. Commented out to save space.
%
Previous work has used social media to detect events, including monitoring disasters \cite{sakaki:2010}, clustering  newsworthy tweets in real-time \cite{mccreadie2013scalable,petrovic2010}, and forecasting popularity of news \cite{bandari:2012}.

Social media has also been used for exploring people's opinions towards objects, individuals, organizations and activities. For example, \newcite{tumasjan:2010} and \newcite{oconnor:2010} have applied sentiment analysis on tweets and predicted election results. \newcite{hu:2004} analyzed restaurant ratings based on online reviews, which contain both subjective and objective sentences. \newcite{golder2011diurnal} and \newcite{dodds:2011} are interested in the temporal changes of mood on social media. \newcite{jmir2534} focus on understanding the perception of emerging tobacco products by analyzing tweets. 

Social media, especially Twitter, has been recently utilized as a popular source of data for public health monitoring, such as
tracking diseases~\cite{ginsberg:2009,yom-tov:2014,nascimento:2014,greene:2011,chew:2010}, mining drug-related adverse events~\cite{bian:2012}, predicting postpartum psychological changes in new mothers \cite{dechoudhury:2013}, and detecting life satisfaction~\cite{schwartz:2013} and obesity~\cite{chunara:2013,cohen-cole:2008,fernandez-luque:2011}.

We focus our attention on the language of food on social media to identify overweight communities and individuals. In the last couple of years, several variants of this problem have been considered~\cite{fried:2014,abbar:2015,culotta:2014,ardehaly:2015}. 
Fried et al.\ \shortcite{fried:2014} collected a large corpus of over three million food-related tweets and use it to predict several population characteristics, namely diabetes rate, overweight rate and political tendency. Generally, they use state-level populations, e.g., one of their classification tasks is to label whether a state is more overweight than the national median.
Overweight rate is the percentage of adults whose Body Mass Index (BMI) is larger than a normal range defined  by NIH. The classification task is to label whether a state is more overweight than the national median.
Individuals' tweets are localized at state level as a single instance to train several classifier models, and the performance of models is evaluated using leave-one-out cross-validation.
Importantly, Fried et al.\ \shortcite{fried:2014} train and test their models on  communities rather than individuals, which limits the applicability of their approach to individualized public health.

Abbar et al.\ \shortcite{abbar:2015} also used aggregated information for predicting obesity and diabetes statistics. They considered energy intake based on caloric values in food mentioned on social media, demographic variables, and social networks. This paper begins to address individual predictions, based on the simplifying assumption that all individuals can be labeled based on the known label of their home county, e.g., all individuals in an  overweight county are overweight, which is less than ideal. In contrast, our work collects actual individual information through the survey derived from community information.

Even though performing classification at state or county granularity tends to be robust and accurate \cite{fried:2014},  characteristics that are specific to individuals are more meaningful and practical.
A wave of computational work on the automatic identification of latent attributes of individuals has recently emerged. Ardehaly and Culotta \ \shortcite{ardehaly:2015} utilize label regularization, a lightly supervised learning method, to infer latent attributes of individuals, such as age and ethnicity. Other efforts have focused on inferring the gender of people on Twitter~\cite{bamman:2014,burger:2011} or their location on the basis of the text in their tweets~\cite{cheng:2010,eisenstein:2010}. These are exciting approaches, but it is unlikely they will perform as well as a fully supervised model, which is the ultimate goal of our work.

%% file: method.tex
\section{Method}

Fried et al.~\shortcite{fried:2014} showed that states and large cities generate a considerable number of food-related tweets, which can be used to infer important information about the respective community, such as 
overweight status or diabetes risk. In an initial experiment, we tested this classifier on the identification of 
overweight {\em individuals}. This classifier did not perform better than chance, likely due to the fact that individuals have a much 
sparser social media presence than entire communities (most tweeters post hundreds of tweets, not millions, and rarely directly about food). 
This convinced us that a realistic public health tool that identifies individuals at risk must be trained on individual data directly, in 
order to learn to take advantage of the specific signal available. 

We describe next the process through which we acquire such data.

\subsection{An interpretable model for community classification}

Our main data-collection idea is to use a playful 20-questions-like survey, automatically generated from a community-based model, which can be 
widely deployed to acquire training data on individuals. 

\begin{figure}[t]
\centering 
\includegraphics[width=0.95\columnwidth]{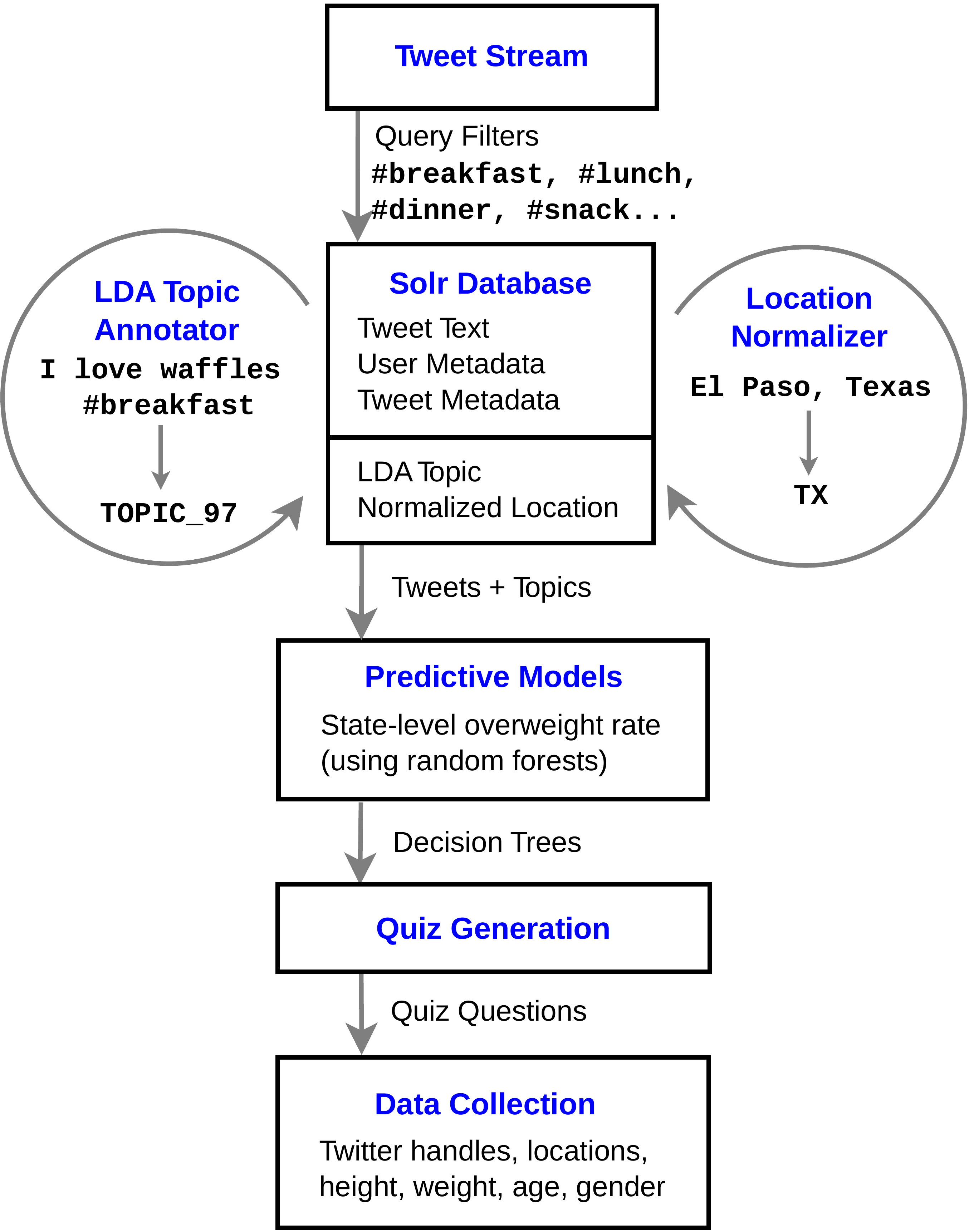}
\caption{Architecture of the semi-automatic approach for quiz generation from social media data.}
\label{fig:arch}
\end{figure}

Our approach is summarized in Figure~\ref{fig:arch}. The first step is to develop an {\it interpretable} predictive model that identifies communities that are more overweight than average, in a way that can be converted into fun, engaging natural language questions. To this end, we started with the 
same settings as \newcite{fried:2014}: we used the 887,310 tweets they collected which were localizable to a specific state and contained at least one relevant hashtag, such as {\tt \#breakfast} or {\tt \#dinner}. Each state 
was assigned a binary label (more or less overweight than the median) by comparing the percentage of overweight adults against the median state. 
For each state, we extracted features based on unigram (i.e., single) words and hashtags from {\em all} the above tweets localized to the corresponding state. To mitigate sparsity, we also included topics generated using Latent 
Dirichlet Allocation (LDA)~\cite{blei:2003} and all tweets collected by Fried et al. For example, one of the generated topics contains words that approximate the standard American diet (e.g., {\em chicken}, {\em potatoes}, {\em cheese}, {\em baked}, {\em beans}, {\em fried}, {\em mac}), which has already been shown to correlate with higher overweight and T2DM rates~\cite{fried:2014}.
% American diet = topic 182 from new_LDA_allTokens.txt

Unlike \newcite{fried:2014}, we do not use support vector machines, but rather a random forest (RF) classifier\footnote{\url{https://code.google.com/p/fast-random-forest/}}. The motivation for this decision was interpretability: as shown 
below, decision trees can be easily converted into a series of {\em if \dots then \dots else \dots} statements, which form the building blocks of the quiz. To minimize the number of questions, we trained a random forest with 7 trees with maximum depth of 3, and we ignored tokens that appear fewer than 3 times in the training data.
These parameter values were selected to make the quiz of reasonable length. We aimed at 20 questions, as in the popular ``20 questions'' game, in which one player must guess what object the other is thinking of by asking 20 or fewer yes-or-no questions. Further tuning confirmed that a small number of shallow trees are most effective in accurately partitioning the state-level data.
% The code used to generate and test this classifier is available at {\small \url{http://github.com/clulab/twitter4food/tree/twitterforest}}.

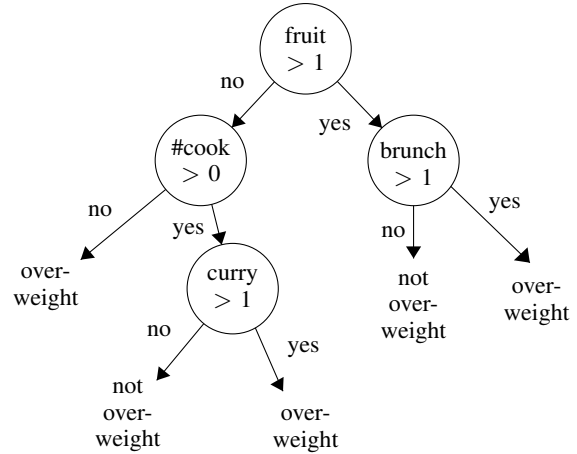
\begin{figure}[t]
\centering 
\begingroup
\fontsize{9pt}{10pt}\selectfont
\begin{tikzpicture}[scale=0.2]
\tikzstyle{every node}+=[inner sep=0pt]
\draw [black] (37.9,-7.1) circle (3);
\draw (37.9,-7.1) node[text width=1cm,align=center] {fruit\\$>$ 1};
\draw [black] (31,-14.5) circle (3);
\draw (31,-14.5) node[text width=1cm,align=center] {\#cook\\$>$ 0};
\draw [black] (45,-14.5) circle (3);
\draw (45,-14.75) node[text width=1cm,align=center] {brunch\\$>$ 1};
\draw (20.8,-23) node[text width=1cm,align=center] {over-\\weight};
\draw [black] (33.1,-23) circle (3);
\draw (33.1,-23) node[text width=1cm,align=center] {curry\\$>$ 1};
\draw (45,-24.2) node[text width=1cm,align=center] {not\\over-\\weight};
\draw (53.1,-24) node[text width=1cm,align=center] {over-\\weight};
\draw (37.9,-32.4) node[text width=1cm,align=center] {over-\\weight};
\draw (26.2,-31.4) node[text width=1cm,align=center] {not\\over-\\weight};
\draw [black] (35.85,-9.29) -- (33.05,-12.31);
\fill [black] (33.05,-12.31) -- (33.96,-12.06) -- (33.23,-11.38);
\draw (33.92,-9.34) node [left] {no};
\draw [black] (39.98,-9.26) -- (42.92,-12.34);
\fill [black] (42.92,-12.34) -- (42.73,-11.41) -- (42.01,-12.1);
\draw (40.92,-12.27) node [left] {yes};
\draw [black] (28.7,-16.42) -- (23.1,-21.08);
\fill [black] (23.1,-21.08) -- (24.04,-20.95) -- (23.4,-20.18);
\draw (24.39,-18.26) node [above] {no};
\draw [black] (31.72,-17.41) -- (32.38,-20.09);
\fill [black] (32.38,-20.09) -- (32.67,-19.19) -- (31.7,-19.43);
\draw (31.29,-19.2) node [left] {yes};
\draw [black] (47.45,-16.22) -- (52.65,-21.28);
\fill [black] (52.65,-21.28) -- (52.50,-20.26) -- (51.6,-21.12);
\draw (51,-18) node [above] {yes};
\draw [black] (45,-17.5) -- (45,-21);
\fill [black] (45,-21) -- (45.5,-20.2) -- (44.5,-20.2);
\draw (44.5,-19.25) node [left] {no};
\draw [black] (34.59,-25.6) -- (36.41,-29.8);
\fill [black] (36.41,-29.8) -- (36.45,-28.85) -- (35.58,-29.35);
\draw (38.84,-27) node [left] {yes};
\draw [black] (31.2,-25.32) -- (28.1,-29.08);
\fill [black] (28.1,-29.08) -- (29,-28.78) -- (28.23,-28.15);
\draw (29.09,-25.77) node [left] {no};
\end{tikzpicture}
\endgroup
\caption{A decision tree from the random forest classifier trained using state-level Twitter data.}
\label{fig:onetree}
\vspace{-3mm}
\end{figure}

To further increase the interpretability of the model, word and hashtag counts were automatically discretized into three bins (e.g.,\ infrequent, somewhat frequent, and very frequent) based on the quantiles of the training data. Figure~\ref{fig:onetree} illustrates one of the decision trees in the trained random forest, with 0 standing for the infrequent, 1 for the somewhat frequent, and 2 for the very frequent bin of the corresponding word or hashtag. 
The figure highlights that the tree is immediately interpretable. For example, the left-most branch indicates that a state is 
classified as overweight if its tweets mention the word ``fruit'' infrequently or somewhat frequently ($fruit \not > 1$), and the hashtag ``\#cook'' 
appears infrequently ($\#cook \not > 0$).\footnote{Despite its simplicity, the proposed RF model performs better than the SVM model of \newcite{fried:2014} on the same data.}
A state with infrequent mention of the word {\it fruit} would take the left branch, then test for the frequency of {\it \#cook}. If 
this is not an infrequent token, then the classifier tests for {\it curry}; very frequent use of {\it curry} would lead to an 
``overweight'' classification (relative to the median state).

\subsection{Quiz}

We next manually converted all decision statements in the random forest classifier into natural language questions. The main assumption 
behind this process is that {\em language use parallels actual behavior}, e.g., 
a person who talks about fruit on social media will also eat fruit in real life. This allowed us to produce more intuitive questions, 
such as {\em How often do you eat fruit?} for the top node in Figure~\ref{fig:onetree}, instead of {\em How often do you mention 
``fruit'' in your tweets?} Table~\ref{tab:questions} shows the questions and corresponding answers we used for the three left-most decision nodes in Figure~\ref{fig:onetree}.
Conversion to natural language questions was as consistent as possible. For example, whenever the relevant feature's word
was a food name {\em x}, the question would be formulated as ``How often do you eat {\em x}?'' with an accompanying picture of 
the food named. When the relevant word was not a food (such as {\em hot} or {\em supper}) or a topic (such as the cluster 
containing {\em diner}, {\em bacon}, {\em omelette}, etc.), the question was formulated in terms of proportion of meals rather than 
frequency.

In all, we generated 33 questions that cover all decision nodes in the random forest classifier. However, when taking the quiz, 
each individual participant answered between 12 and 24 questions, depending on their answers and the corresponding traversal of 
the decision trees.
%To improve the user experience, the features used were food words and hashtags, 
% as well as LDA topics, as in~\cite{fried:2014}. 

\begin{table}[t]
\begin{tabular}{p{7cm}}
\toprule
How often do you eat fruit? \\
\hspace{2mm} $\rightarrow$ {\it Practically never, Sometimes, Often}	\\
What proportion of your meals are home cooked? \\
\hspace{2mm} $\rightarrow$ {\it None or very little, About half, Most or all}	\\
How often do you eat curry? \\
\hspace{2mm} $\rightarrow$ {\it Practically never, Sometimes, Often} \\
\bottomrule
\end{tabular}
\caption{Example questions derived from the decision nodes in Figure~\ref{fig:onetree}.}\label{tab:questions}
\vspace{-4mm}
\end{table}

\begin{figure*}
	\centering
	\begin{subfigure}{0.4\textwidth}
		\includegraphics[scale=.75]{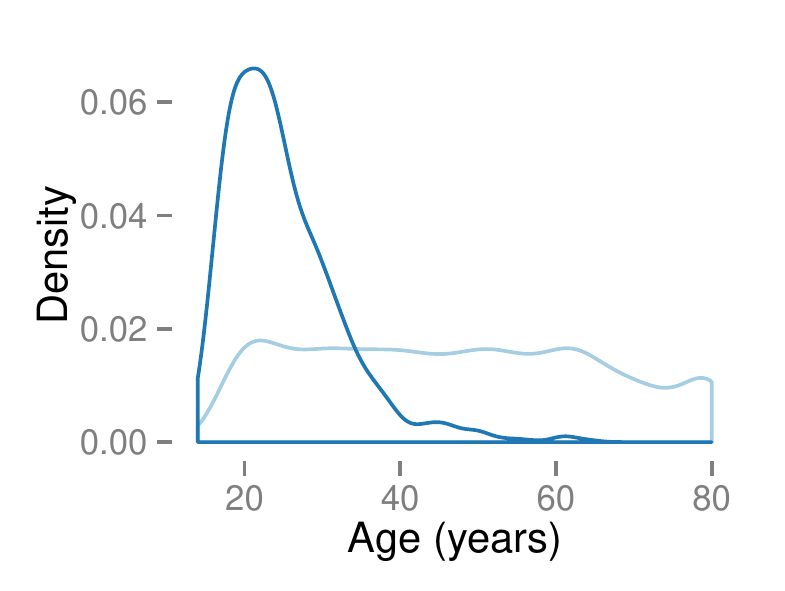}
		\caption{Age distribution in the NHANES survey and in the present study's online quiz.}
	\end{subfigure} \hspace{1em}
	\begin{subfigure}{0.5\textwidth}
		\includegraphics[scale=.75]{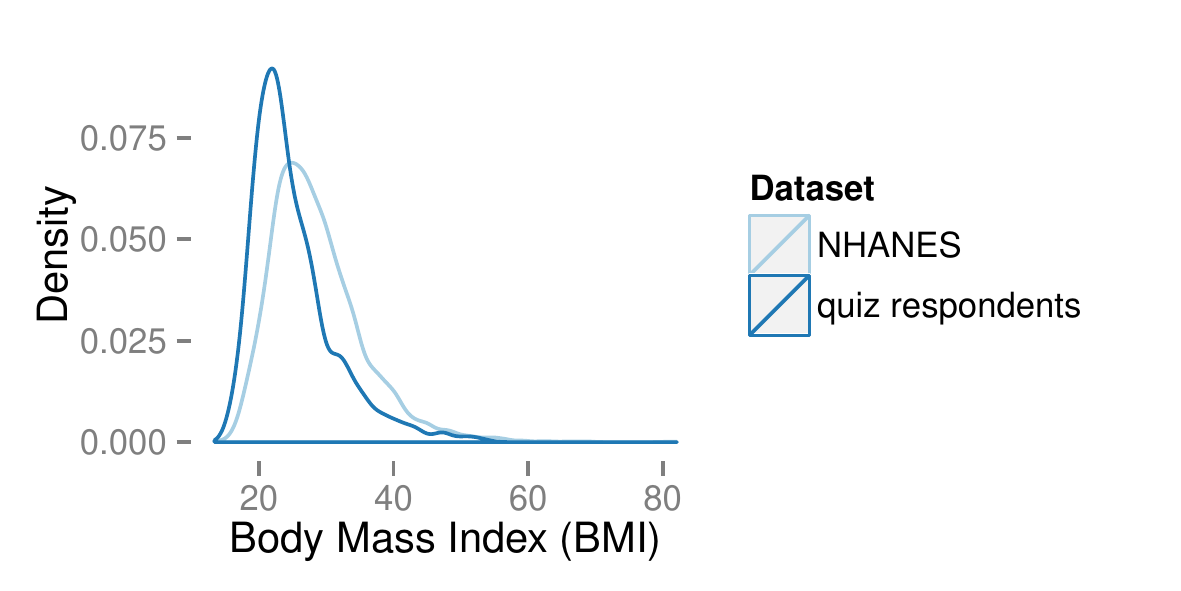}
		\caption{BMI distribution in the NHANES survey and in the present study's online quiz.}
	\end{subfigure}

	\begin{subfigure}{0.45\textwidth}
		\includegraphics[scale=.75]{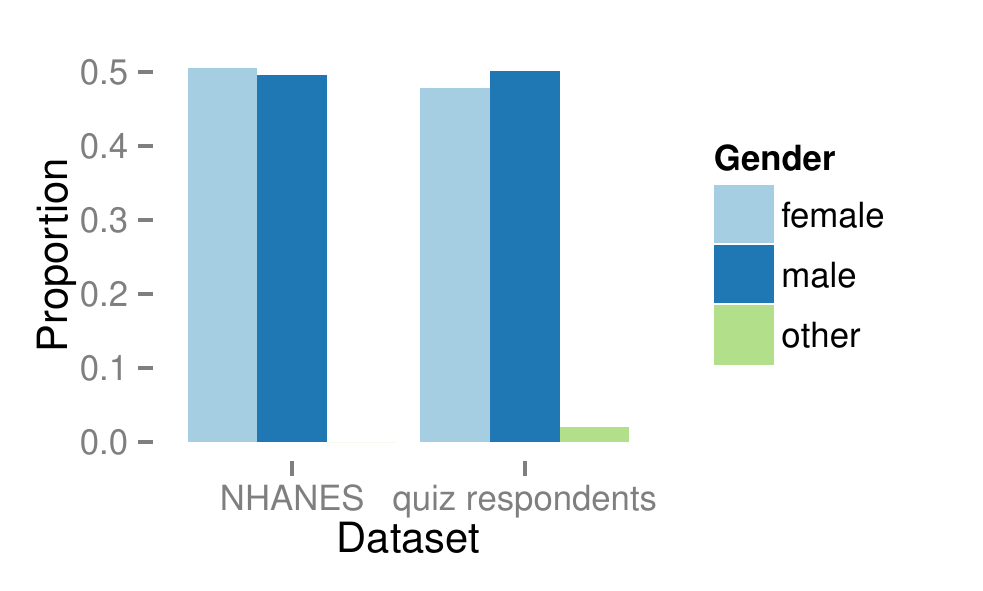}
		\caption{Gender distribution in the NHANES survey and in the present study's online quiz.}
	\end{subfigure} \hspace{1em}
	\begin{subfigure}{0.45\textwidth}
		\includegraphics[scale=.75]{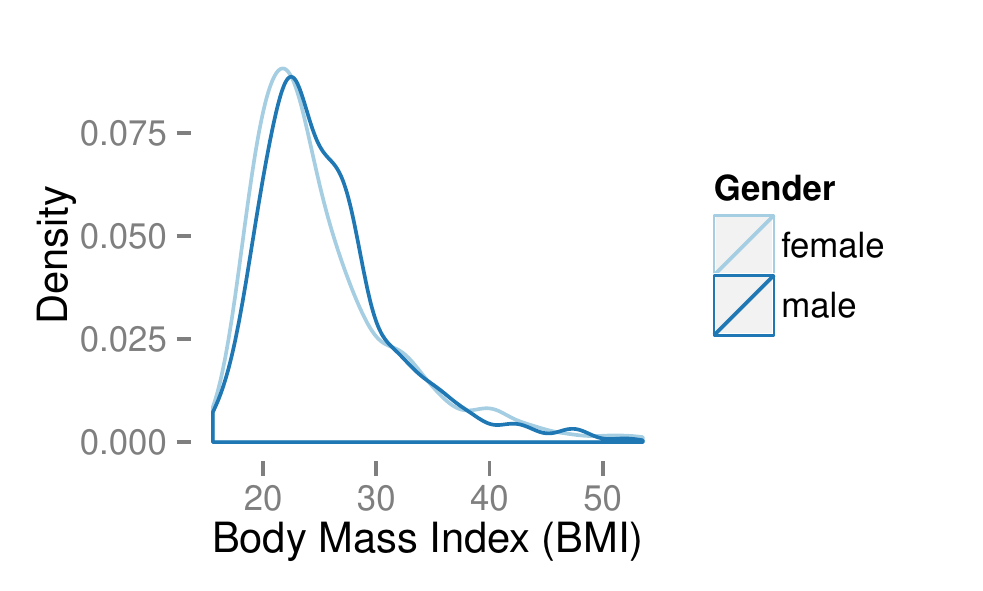}
		\caption{BMI distribution by gender in the present study. Too few respondents marked a gender of {\em other} to allow a density plot.}
	\end{subfigure}
	\begin{subfigure}{0.45\textwidth}
		\includegraphics[trim={2cm 0 2cm 0},clip,scale=.75]{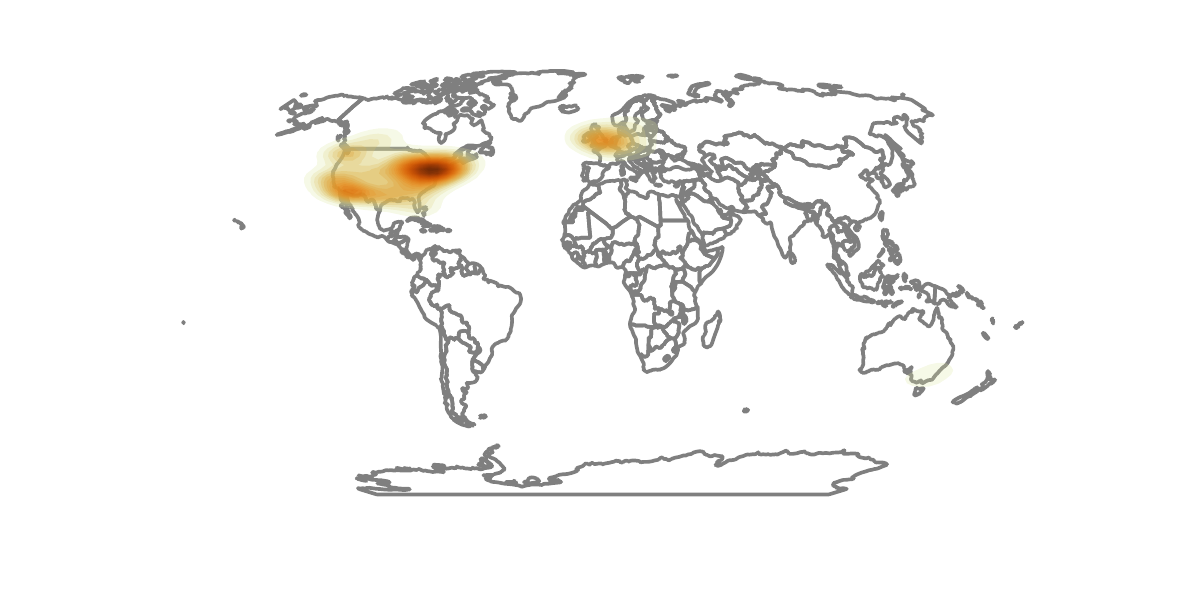}
		\caption{Geographical distribution of survey respondents worldwide. Of 625 respondents providing location information, 421 (67.4\%) provided a location within the US.}
	\end{subfigure} \hspace{1em}
	\begin{subfigure}{0.45\textwidth}
		\includegraphics[trim={2cm 0 1.5cm 0.5cm},clip,scale=.75]{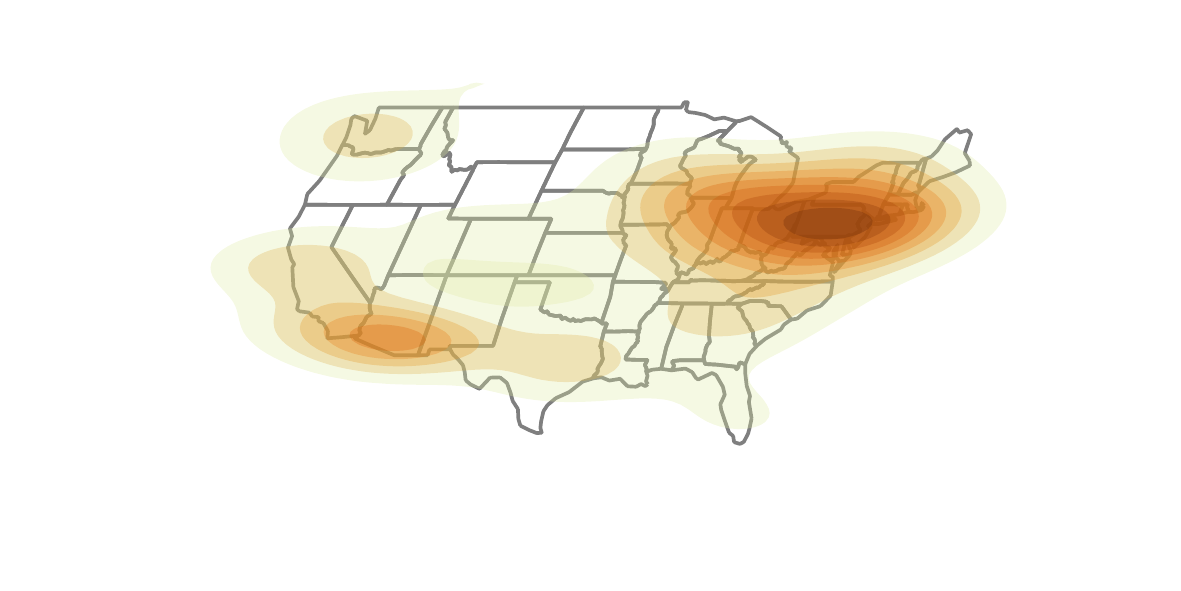}
		\caption{Geographical distribution of survey respondents in the US.}
	\end{subfigure}
	\caption{Demographic data from the present study and from the National Health and Nutrition Examination Survey (NHANES) \cite{nhanes2013}. Of 945 respondents, 833 provided their age, 864 provided information for their BMI calculation, 846 provided their gender, and 625 provided their location.}
	\label{fig:demo}
\end{figure*}

This quiz serves to gather training data, which will be used in future work to train a supervised model for the identification 
of individuals at risk. 
To our knowledge, this approach is a novel strategy for quiz generation, and it serves as an important stepping-stone toward our goal of building individualized 
public health tools driven by social media. 
With respect to data retention, we collect (with the permission of the participants) the following additional data to be used for future research: 
height, weight, sex, location, age, and social media handles for Twitter, Instagram, and Facebook. 
We only downloaded public posts using these handles.
This data (specifically height and weight) is also immediately used to compute the participant's BMI, to verify whether the 
classifier was correct.

%% file: results.tex
\section{Empirical results}

\subsection{Evaluation of random forest classifier}

\begin{table}[t]
\centering
\begin{tabular}{l l}
\toprule
{\it Model}				& {\it Accuracy}	\\
\midrule
Majority baseline		& 50.89		\\
SVM (Fried et al., 2014)	& 80.39		\\
%This is the SVM performance trained on several months more of data. Include?
%SVM (food words + hashtags + LDA)	& 86.27		\\
RF (food + hashtags)					& 82.35	\\
Discretized RF	 (food + hashtags)		& 78.43		\\
\bottomrule
\end{tabular}
\caption{Random forest (RF) classifier performance on state-level data relative to majority baseline and Fried et al. (2014)'s best classifier. We include two versions of our classifier: the first keeps numeric features (e.g., word counts) as is, whereas the second discretizes numeric features to three bins.}
\label{tab:rf}
% \vspace{-4mm}
\end{table}

Table~\ref{tab:rf} lists the results of our RF classifier on the task of classifying overweight/not-overweight states. We used identical experimental settings as~\cite{fried:2014}, i.e., leave-one-out-cross-validation on the 50 states plus the District of Columbia. The table shows that our best model performs 2\% better than the best model of~\cite{fried:2014}. Our second classifier, which used discretized numeric features and was the source of the quiz, performed 2\% worse, but it still had acceptable accuracy, nearing 80\%. 
As discussed earlier, this discretization step was necessary to create intelligible Likert-scaled questions~\cite{likert1932scale}.

%\todo{I do not understand this paragraph! These numbers, especially 84.31, do not match the data in table. Dane, let's discuss.}
%Decision trees can perform accurate classification, and have the advantage of being highly interpretable. In fact, the random forest classifier trained with these parameters outperformed (with more data) \cite{fried:2014} on by-state classification of weight of 84.31 compared to 80.39 without discretizing the features. Discretization lowered accuracy somewhat to 78.43, but this step was necessary to create intelligible Likert-scaled questions.

\subsection{Quiz response}

\begin{figure}
\centering
\begin{subfigure}[b]{\columnwidth}
	\includegraphics[width=\textwidth]{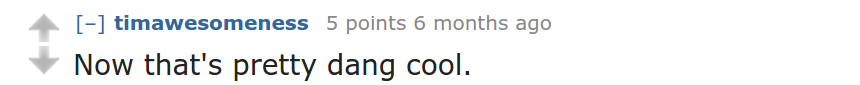}
	\caption{affective}
\end{subfigure}\vspace{1em}

\begin{subfigure}[b]{\columnwidth}
	\includegraphics[width=\textwidth]{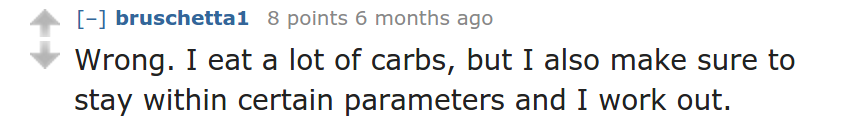}
	\caption{hypothesizing}
\end{subfigure}\vspace{1em}

\begin{subfigure}[b]{\columnwidth}
	\includegraphics[width=\textwidth]{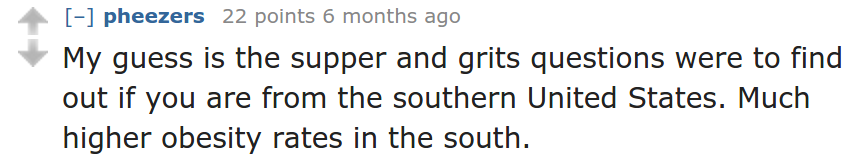}
	\caption{cultural}
\end{subfigure}\vspace{1em}

\begin{subfigure}[b]{\columnwidth}
	\includegraphics[width=\textwidth]{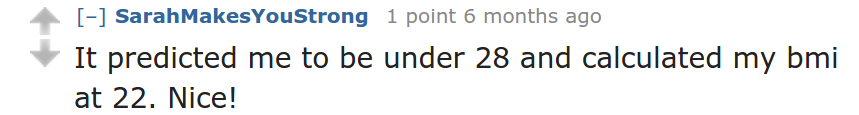}
	\caption{result-based}
\end{subfigure}\vspace{1em}

\begin{subfigure}[b]{\columnwidth}
	\includegraphics[width=\textwidth]{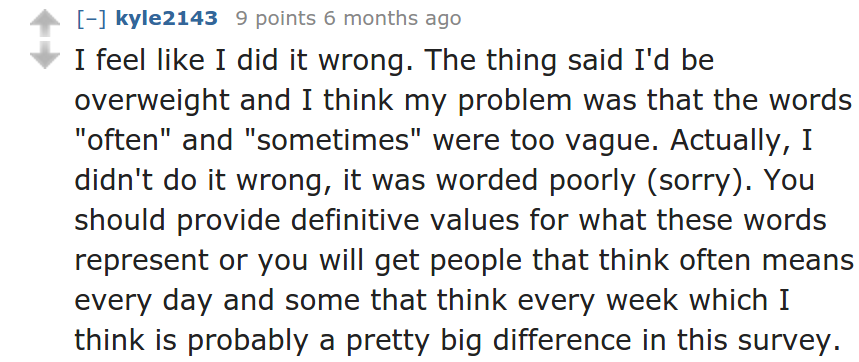}
	\caption{constructive criticism}
\end{subfigure}\vspace{1em}

\begin{subfigure}[b]{\columnwidth}
	\includegraphics[width=\textwidth]{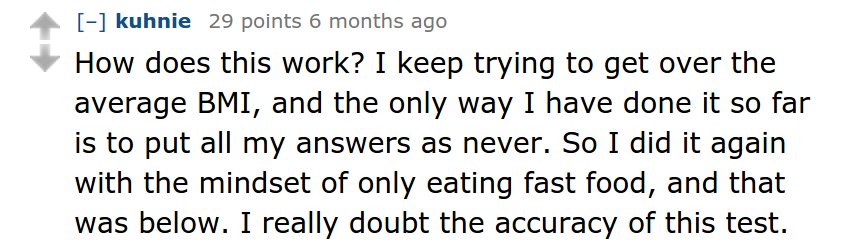}
	\caption{understanding}
\end{subfigure}
\caption{Some comments of various classes from Redditors in response to the quiz.}\label{fig:redditcomments}
\end{figure}

\begin{table}
\centering
\begin{tabular}{@{} p{2cm} l p{5cm} @{}}
\toprule
{\it Comment type}		& \%		& {\it Example} \\ \midrule
affective	& 11	 & This is awesome. Good luck and keep up the great work!	\\[0.2em]
hypothesizing	& 20& I probably stumped your system because I love all types of food, but I know how to portion correctly.	\\[0.1em]
cultural	& 5	& was the question about ``supper'' meant to isolate people from the Midwest?		\\[0.2em]
result-based	& 13	& Surprised that this was correct, I eat like a fat person. Good job!	\\[0.2em]
demographic	& 25	& Single, Caucasian, eat fast food 3-5 times a week	\\[0.2em]
constructive criticism	& 26	& Further breaking down the options would be better.	\\
\bottomrule
\end{tabular}
\caption{Comments submitted with the demographic questionnaire on the quiz site.}\label{tab:quizcomments}.
\vspace*{-3mm}
\end{table}

Many of the 945 participants were highly engaged with the quiz; 97.9\% volunteered demographic information at the end of the quiz. Many of the participants also left feedback, some on the Reddit page linking to the quiz, as shown in Figure \ref{fig:redditcomments}, some on the quiz page itself, as shown in Table \ref{tab:quizcomments}. The feedback comprised mostly comments on the accuracy (or inaccuracy) of the quiz, comments expressing interest in particular questions, and speculation about how the quiz was constructed.

It seems that quiz accuracy was not a prerequisite for commenting on the quiz. On the contrary, participants were more likely to comment when their results were inaccurate. It is unknown whether the up- and down-voting was motivated by the accuracy of the quiz, but researchers making interactive prediction sites may discover that inaccuracy is in fact more engaging in some regards. The perceived stigma of obesity was also evident in the reactions to the quiz, with some negative reactions to a prediction of overweight regardless of its accuracy.

%Some participants reported that the nonspecificity of the choices made the task difficult, because they had to judge whether their consumption of a food or use of a food word was rare of common compared to an unknown norm instead of a specific rate such as, e.g., ``once per week''.

For a better understanding of the feedback received, we performed a post-hoc analysis. Our analysis indicated that while there were 3 comments made about accuracy out of 744 people with correct predictions (0.40\% commented), and 13 noting incorrect answers out of 201 with incorrect predictions (6.5\% commented). Thus, the participants were 16 times as likely to comment on the quiz's accuracy if its prediction was incorrect than they were if it was correct. We further classified the Reddit comments received into six classes: affective comments (7\%), hypothesizing comments (17\%), cultural comments (17\%), result-based comments (53\%), constructive criticism (7\%), and comments seeking a greater understanding of the quiz (7\%)\footnote{Numbers sum to more than 100\% because of comments with multiple classes of content.}, examples of which can be seen in Figure~\ref{fig:redditcomments}. The comments made on the quiz site also frequently included additional diet and demographic information.

\subsection{Quiz evaluation}

We evaluated the quiz on 945 volunteers recruited at the University of Arizona and on social media, namely Facebook, Twitter, and Reddit's SampleSize subreddit\footnote{http://www.reddit.com/r/SampleSize/}. The results are summarized in Table~\ref{tab:results}. We evaluated the accuracy of the random forest classifier by comparing each individual's actual BMI, based on the self-reported height and weight, to the classifier's prediction. The cutoff boundary BMI for both training and testing was 28.7 -- the average US adult BMI according to \newcite{nhanes2013}. This figure is above the NIH's definition of {\it overweight} (BMI $\geq$ 25) because the average US resident is overweight by that standard.
These results are promising: the quiz had a 78.7\% accuracy for the classification of individuals into the two classes: higher or lower BMI than the average US resident.

\begin{table}
\begin{tabular}{@{} l l @{}}
%Statistics	& Value	\\
\toprule
Quiz accuracy				& 78.7\%				\\
Accuracy, participants BMI $\geq$ 28.7 	& 16.0\%\\
Accuracy, participants BMI $<$ 28.7 	& 92.2\%\\
\midrule
Proportion, participants BMI $\geq$ 28.7 	& 17.7\%				\\
Proportion, participants BMI $<$ 28.7 	& 82.3\%				\\
\midrule
Mean participant weight		& 74.4 kg (164 lbs)		\\
Mean US adult weight	& 80.3 kg (177 lbs)		\\
\midrule
Mean participant height		& 173 cm (5 ft 8 in)	\\
Mean US adult height	& 167 cm (5 ft 6 in) 	\\
\midrule
Mean participant BMI		& 24.9				\\
Mean US adult BMI	& 28.6				\\
\midrule
Mean participant age		& 26.1 years			\\
Mean US adult age	& 47.1 years			\\
\bottomrule
\end{tabular}
\caption{Results of the quiz evaluation, together with statistics of the adult participants (18 or older) who took the test, compared against average values in the US~\cite{cdc14diagnosed}.}
\label{tab:results}
\end{table}

%% file: discussion.tex
\section{Discussion}

It is important to note that the limitations of the sample are considerable: as shown in Table~\ref{tab:results} and Figure~\ref{fig:demo}, 
our initial sample is taller, lighter, and younger than the average US adult, leading to a biased test sample. The strong bias of the sample means that the trivial baseline of predicting no participants to be over BMI 28.7 would have accuracy 82.3\%. 

Moreover, while the overall accuracy of the random forest backing the quiz was 78.7\%, the accuracy on participants who reported a BMI over 
28.7 was only 16.0\%. Our conjecture is that, in general, participants who are overweight are more reluctant to mention food- and health-related topics on social media, which led to lower-quality training data for this group, and the distribution of BMI for participants classified as under 28.7 was not significantly different from that of those classified as over 28.7.

Far from being a problem for this data collection technique, however, the failure of transfer from state-level training to individual-level testing underscores the need for the data collection itself. No existing system has been able to automatically predict individuals' weight after training on state-level data.

%% file: conclusion.tex
\section{Conclusions and future work}

We described a strategy for the acquisition of training data necessary to build a social-media-driven early detection system for individuals at risk for T2DM, using a game-like quiz with data and questions acquired from Twitter. 
Our approach has proven to inspire considerable participant engagement and, in so doing, 
provide relevant data to train a public-health model applied to individuals.

First, we built  a random forest classifier that improves on the state of the art for the classification of overweight communities (in particular US states). We then use this as the basis of a 20-questions-style quiz to classify individuals. Early results are promising:  78.7\% accuracy, but the sample does not represent the general population well, and the quiz performs poorly on classifying overweight individuals. 

The most immediate goal is to obtain a large respondent sample that is more representative of US adults, and to extend the information gathered to longitudinal data. Based on the high engagement observed in this initial experiment, we hope that a large dataset can be constructed at minimal cost. This dataset will be used to develop a public-health tool capable of non-intrusively identifying at-risk individuals by monitoring public social-media streams.

Our long term goal is to use this data to train a supervised classifier for the identification of individuals at risk for type 2 diabetes. The dataset collected through the quiz described here is sufficient for this goal: it includes necessary information for the calculation of BMI (weight, height), demographic information, and social media handles. We plan to explore (public) multi-modal social media information: natural language, posted pictures, etc. From this, we will extract and use preventable risk factors, such as poor diet or lack and perceived lack of physical activity. 
The data will be made available to interested researchers.
There is great potential for further improvements of the model by adding calorie count estimates for food pictures associated with individual tweets,  by incorporating
individual-level demographic features such as gender and age, and by using words and hashtags about physical activities.

%% file: resources.tex
\section{Resources}

\begin{itemize}
\item The software used to generate and test the random forest classifier is open-source at \url{http://github.com/clulab/twitter4food/}
\item The quiz is available from the project's main page at \url{http://sites.google.com/site/twitter4food/}
\item Anonymized quiz results are available at \url{http://git.io/vZY5U}. They detail the responses of each participant to the quiz questions, the system's prediction, its accuracy, and the participants' height, weight, location, age, gender, and (anonymized) comments.
\end{itemize}